\documentclass[a4paper,UKenglish,cleveref, autoref, thm-restate]{lipics-v2021}

\nolinenumbers
\hideLIPIcs

\usepackage{xcolor}
\usepackage{xspace}
\usepackage{hyperref}
\usepackage{algorithm} 
\usepackage[noend]{algpseudocode}

\newcommand{\minizinc}[0]{\textsc{MiniZinc}\xspace}
\newcommand{\autoig}[0]{\textsc{AutoIG}\xspace}
\newcommand{\code}[1]{{\small\texttt{#1}}}

\newcommand{\ortools}[0]{OR-Tools\xspace}
\newcommand{\picat}[0]{Picat-SAT\xspace}
\newcommand{\chuffed}[0]{Chuffed\xspace}
\newcommand{\yuck}[0]{Yuck\xspace}

\newcommand{\essence}[0]{\textsc{essence}\xspace}

\newcommand{\conjure}[0]{\textsc{conjure}\xspace}
\newcommand{\savilerow}[0]{\textsc{savilerow}\xspace}
\newcommand{\minion}[0]{\textsc{minion}\xspace}
\newcommand{\irace}[0]{\textsc{irace}\xspace}
\newcommand{\macc}[0]{\textsc{macc}\xspace}
\newcommand{\carpetcutting}[0]{\textsc{carpet-cutting}\xspace}
\newcommand{\mario}[0]{\textsc{mario}\xspace}
\newcommand{\racp}[0]{\textsc{racp}\xspace}
\newcommand{\lotsizing}[0]{\textsc{lot-sizing}\xspace}
\newcommand{\runsolver}[0]{\textsc{runsolver}\xspace}

\usepackage[utf8]{inputenc}
\usepackage[T1]{fontenc}
\usepackage{listings}
\usepackage{courier}
\definecolor{mygreen}{rgb}{0,0.6,0}
\definecolor{myblue}{rgb}{0,0,0.6}
\definecolor{myred}{rgb}{0.6,0,0}
\lstdefinelanguage{essence}{
keywordstyle=\color{myblue}\bfseries,
keywordstyle=[2]\color{teal}\bfseries,
keywordstyle=[3]\color{mygreen}\bfseries,
keywordstyle=[4]\color{myred}\bfseries,
identifierstyle=\color{black},
commentstyle=\color{purple}\ttfamily,
stringstyle=\color{red}\ttfamily,
keywords = {language, Essence, given, letting, find,
            domain, total, surjective, be,
            injective, in, preImage, range,
            new, type, intersect, union, from,
            of, indexed, by, and, or, toInt, numParts, partSize, together, 
            defined, maxSize, maxNumParts, size, regular, language, cheese, ->
            },
morekeywords=[2]{int, bool, set, mset, variant, record, sequence, function, relation, partition,matrix, enum},
morekeywords=[3]{such, that, minimising, maximising},
morekeywords=[4]{forAll, exists, sum},
otherkeywords={:,=,|},
sensitive=false,
comment=[l]{\%},
morestring=[b]',
morestring=[b]",
}

\title{A Framework for Generating Informative Benchmark Instances}

\author{Nguyen Dang}{School of Computer Science, University of St Andrews,
  United Kingdom}{nttd@st-andrews.ac.uk}{
  https://orcid.org/0000-0002-2693-6953}{is a Leverhulme Early Career Fellow}

\author{Özgür Akgün}{School of Computer Science, University of St Andrews,
  United Kingdom}{ozgur.akgun@st-andrews.ac.uk}{https://orcid.org/0000-0001-9519-938X}{}

\author{Joan Espasa}
{School of Computer Science, University of St Andrews, United Kingdom}
{jea20@st-andrews.ac.uk}
{https://orcid.org/0000-0002-9021-3047}
{}

\author{Ian Miguel}{School of Computer Science, University of St Andrews,
  United Kingdom}{ijm@st-andrews.ac.uk}{
  https://orcid.org/0000-0002-6930-2686}{supported by EPSRC EP/V027182/1}

\author{Peter Nightingale}{Department of Computer Science, University of York,
  United Kingdom}{peter.nightingale@york.ac.uk}{
  https://orcid.org/0000-0002-5052-8634}{}

\authorrunning{N. Dang, Ö. Akgün, J. Espasa, I. Miguel, P. Nightingale}

\Copyright{Nguyen Dang, Özgür Akgün, Joan Espasa, Ian Miguel, Peter Nightingale}



\supplementdetails[]{Code}{https://github.com/stacs-cp/AutoIG} 

\acknowledgements{This work uses the Cirrus UK National Tier-2 HPC Service at EPCC (\url{http://www.cirrus.ac.uk}) funded by the University of Edinburgh and EPSRC (EP/P020267/1).}

\EventEditors{Christine Solnon}
\EventNoEds{1}
\EventLongTitle{28th International Conference on Principles and Practice of Constraint Programming (CP 2022)}
\EventShortTitle{CP 2022}
\EventAcronym{CP}
\EventYear{2022}
\EventDate{July 31--August 8, 2022}
\EventLocation{Haifa, Israel}
\EventLogo{}
\SeriesVolume{235}
\ArticleNo{26}

\begin{document}

\maketitle

\begin{abstract}

Benchmarking is an important tool for assessing the relative performance of alternative solving approaches.
However, the utility of benchmarking is limited by the quantity and quality of the available problem instances.
Modern constraint programming languages typically allow the specification of a class-level model that is parameterised over instance data.
This separation presents an opportunity for automated approaches to generate instance data that define instances that are \textit{graded} (solvable at a certain difficulty level for a solver) or can \textit{discriminate} between two solving approaches.
In this paper, we introduce a framework that combines these two properties to generate a large number of benchmark instances, purposely generated for effective and informative benchmarking.
We use five problems that were used in the MiniZinc competition to demonstrate the usage of our framework.
In addition to producing a ranking among solvers, our framework gives a broader understanding of the behaviour of each solver for the whole instance space; for example by finding subsets of instances where the solver performance significantly varies from its average performance.



\end{abstract}

\section{Introduction}

A practitioner faced with solving a new problem has a difficult choice among many solving algorithms, whose performance on the new problem is unknown and is likely to be variable. One approach is to draw instances from the problem to {\em benchmark} the various solvers under consideration, i.e. an empirical study of relative performance. This approach is  favoured for computationally challenging tasks since the performance behaviour of a non-trivial algorithm is difficult to predict and is unlikely to be susceptible to a purely theoretical analysis~\cite{bartz2020benchmarking}. As Beiranvand et al.~\cite{beiranvand2017best} argue, care must be taken to select an instance set with a variety of difficulty for benchmarking in order to obtain the best insight into solver performance.

Constraint programming (CP) approaches particularly benefit from empirical analysis, since modern tool chains like \minizinc~\cite{nethercote2007minizinc} and \savilerow~\cite{nightingale2017automatically} support targeting multiple solvers from a  solver-independent constraint model. These may be entirely different paradigms, such as SAT~\cite{biere2009handbook}, SMT~\cite{barrett2018satisfiability} or indeed CP, and so can vary in performance  significantly.



The need for empirical benchmarking is further supported by competitions run by several research communities, like the \minizinc challenge~\cite{stuckey2010philosophy} in the CP community, the SAT competition~\cite{froleyks2021sat} and the AI planning competition~\cite{ipc}.
Solver developers enter a competition by providing a default configuration of their solver. Each solver supports a common interface for specifying their input and output. The competition is then run on a set of problem instances and the solvers are ranked with respect to their comparative performance.

In the main solver competition for CP, the \minizinc challenge, each solver is given two inputs: a solver-independent problem-level model and instance data written in a separate data file. Then \minizinc is used to instantiate and translate the solver-independent model into input suitable for each solver. The main result of the challenge is a ranking of solvers. More detailed results pertaining to the ranking of solvers per problem class are also published.

The selection of problem instances to be used in a competition is extremely important to avoid conclusions that are unintentionally biased towards the chosen instances. Competitions somewhat mitigate this problem by inviting solver authors to submit benchmark instances. This is a promising sociotechnical attempt at alleviating the problem of bias, but it is laborious and does not provide a comprehensive solution.

Benchmarking is not only useful for finding an overall ranking among options, but also for finding subsets of instances where the performance of a solver is significantly different from the performance of the same solver overall. For example, solver \textit{A} might perform better for most instances of a problem class in comparison with \textit{B}, yet perform very poorly for a particular subset of the instances. Information like this can be extremely valuable to solver developers. A traditional competition that works by running all solvers on a fixed set of instances can occasionally detect such cases even though it does not actively look for them.

\begin{figure}[t!]
    \centering
    \includegraphics[width=\linewidth,clip]{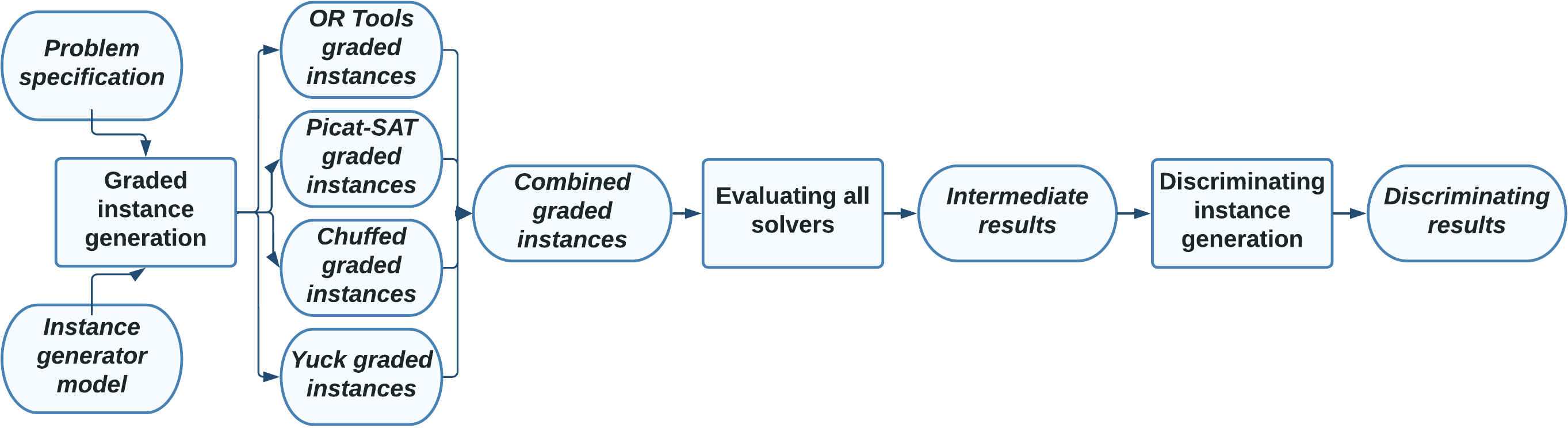}
    \caption{Flowchart of the whole \autoig application process}
    \label{fig:flowchart}
\end{figure}

For an informative benchmark we need a sufficient \textit{quantity} of high \textit{quality} instances and the ability to dynamically explore subsets of the instance space to detect performance discrimination.
In this work we present \autoig, a constraint-based instance generation framework, that supports automatically generating \textit{graded} instances (i.e., solvable at a certain difficulty level for a given solver), and finding discriminating instances (i.e. easy for one solver and difficult for another solver).
In combination, these two methods can be used to generate a large number of high-quality instances. Furthermore, they can be used to find interesting subsets of the instance space 
as opposed to leaving their discovery to chance.

\autoref{fig:flowchart} gives a flowchart for an end-to-end application of \autoig, whose instance generation process is explained in \autoref{sec:autoig_system_description}. Without loss of generality, the flowchart lists the four solvers used for the evaluation of \autoig in this paper. \autoref{sec:case_studies} explains the choice of these solvers and the five problem classes we use.
Both stages of \autoig can be applied to other solvers and solver configurations. 
The \autoig process has two main inputs: a problem specification (in the form of a \minizinc model in this paper) and a problem specific instance generator. The instance generator is parameterised to allow \autoig to generate a variety of instances. There are two main places where we can extract results from \autoig, evaluating all solvers on the combined set of graded instances (marked intermediate results in the flowchart, see \autoref{sec:results_graded}) and evaluating the results of discriminating instances (marked discriminating results in the flowchart, see \autoref{sec:results_discriminating}). \autoig source code and all data and models used in this paper are available at \url{https://github.com/stacs-cp/AutoIG}.



The main contributions of this paper include:
\begin{enumerate}
    \item A novel constraint-based framework for generating informative benchmark instances which combines two approaches (\textit{graded} and \textit{discriminating} instance generation) that were previously used in isolation~\cite{akgun2019instance,akgun2020discriminating}. 
    \item Support for \minizinc and hand-written instance generators. The new system accepts a user-defined generator as a constraint model, thus allowing problem-specific knowledge to be injected into the instance generation process.
    \item Support for the evaluation of local search solvers in addition to systematic solvers. The instance evaluation also considers both solution quality and running time.
    \item An extensive evaluation on five problems from the \minizinc challenge, showing that we can gain new interesting insights that complement the competition's results.
\end{enumerate}

\section{Related Work}
\label{sec:related_work}

A series of papers uses evolutionary algorithms and applies \textit{instance space analysis} methods to problems in machine learning (classification~\cite{munoz2018instance}, regression~\cite{munoz2021instance}, clustering~\cite{fernandes2021towards}) and in combinatorial optimisation (personnel scheduling~\cite{kletzander2021instance}, bin packing~\cite{liu2020using}, course timetabling~\cite{de2021algorithm}). 
They use evolutionary algorithms to generate problem instances~\cite{smith2011discovering, smith2021revisiting}, whereas we take a constraint-based approach. Part of their work is  analysing existing instances in benchmark suites and visualising the hardness distribution of instances for particular problems; our framework can be fruitfully combined with their detailed analysis and visualisation methods.

Instance generators have been applied to hard problems in Operations Research as well. For example, NSPLib~\cite{vanhoucke2009characterization} provides an instance generator and large sets of nurse rostering instances. Their instance generator characterizes an instance through various complexity indicators, including problem sizes, preference distribution measures, coverage distribution measures, and time related constraints. They implement a  dedicated procedure for generating instances with properties corresponding to the values of specific indicators as parameters.
For the knapsack problem, \cite{pisinger2005hard} uses instance generators to identify the regions of the instance space that contain difficult instances. For the traveling thieves problem, \cite{bossek2021generating} uses instance generators that discriminate between more than two options simultaneously.

In communities such as SAT, there have been various works~\cite{sat1,sat2} that try to address the generation of instances with desired properties. The SAT competition~\cite{satcompetition} organisers partly crowdsource the creation of the evaluation set. They require participants to send 20 new instances each, guaranteeing that the competition is run on instances mostly unseen to the solver developers prior to the competition. In addition, a set of previously used instances is manually and carefully selected, using various criteria such as hardness and variety. 

The problem of generating a good set of benchmark instances is also studied in the AI planning community~\cite{torralba2021automatic}. SMAC~\cite{smac}, a tool for optimizing algorithm parameters, is paired with hand-coded programs to generate many sets of instances that smoothly scale in difficulty. Afterwards, a subset of the generated sets is selected, according to various criteria such as difficulty and fairness. This results in a set of instances that better reflect the differences between planners when compared to the instances used in the competition.

A related field of study is algorithm configuration/selection, including portfolio-based approaches (SATZilla~\cite{xu2008satzilla,xu2012satzilla2012}, CPHydra~\cite{o2008using}, sunny-CP~\cite{sunny2015,sunny2021}). For these purposes it is important to have a sufficient number of instances with a variety of difficulties that can discriminate between the options~\cite{schneider2012quantifying}.

\section{Constraint-based Automated Instance Generation}
\label{sec:autoig_system_description}

Following the approaches in~\cite{akgun2019instance} and~\cite{akgun2020discriminating}, our instance generation system \autoig makes use of the \essence constraint modelling pipeline~\cite{essencepipeline} 
and the automated algorithm configurator \irace~\cite{lopez2016irace}. The system receives as input a problem description model, a parameterised instance generator written as a constraint model (referred to as the \emph{generator model}), the solver(s) for which we want to generate graded or discriminating instances, and the types of instances we are interested in (SAT or UNSAT or both). The role of the \essence pipeline is to express the generator model and to create \emph{candidate instances} by solving instances of the generator model (referred to as \emph{generator instances}), while the role of \irace is to search in the parameter space of the generator model, or in other words, to sample in the generator instance space, to find configurations that can give us candidate instances with the desired properties. In this section, we first describe the search procedure of \irace (Section~\ref{sec:irace}). We then explain how \irace and constraint modelling are combined in the instance generation process of \autoig (Section~\ref{sec:autoig}). Finally, we discuss in detail how each candidate instance is evaluated during \autoig search using gradedness or discriminating criteria (Section~\ref{sec:instance_evaluation}).


      


\subsection{\irace's Tuning Process}
\label{sec:irace}

\begin{figure}[t]
    \centering
    \includegraphics[width=0.65\linewidth]{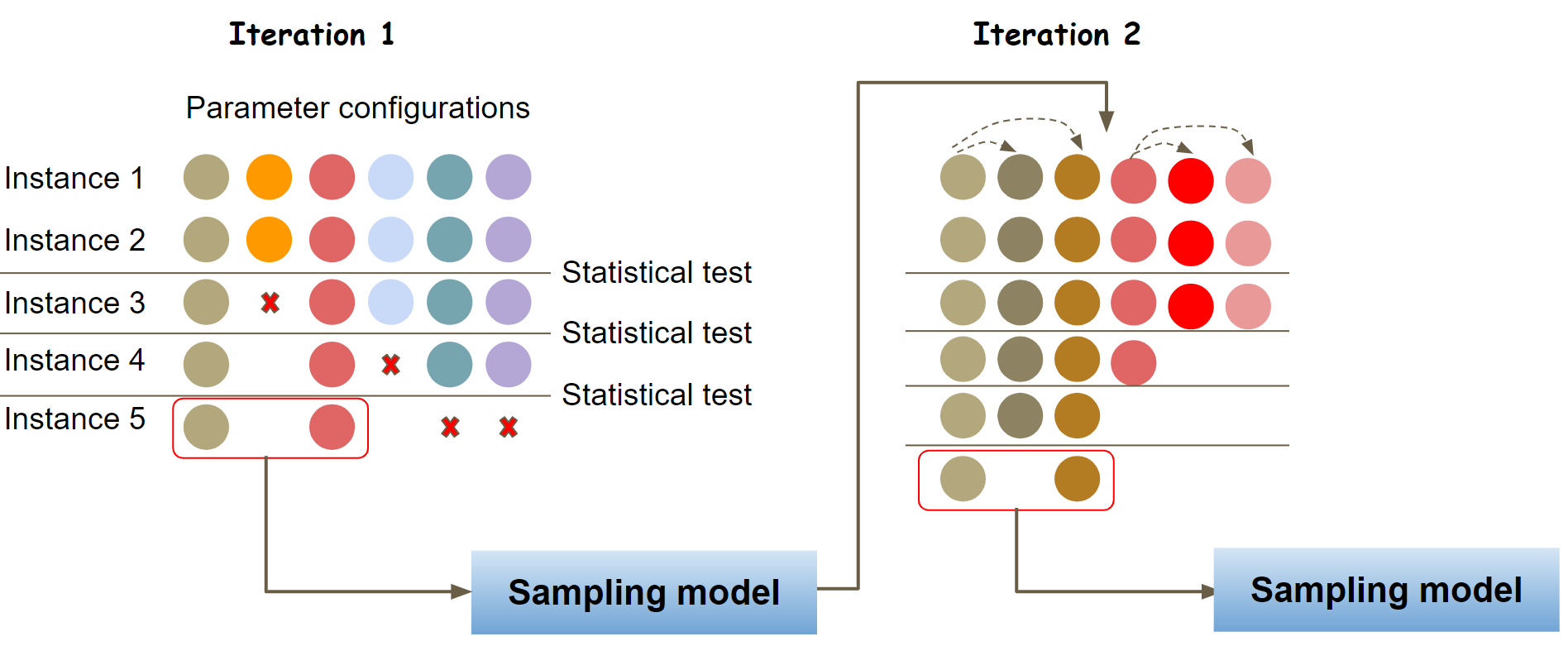}
    \caption{An illustration of \irace's tuning process.}
    \label{fig:irace}
\end{figure}

\irace~\cite{lopez2016irace} is a general-purpose automated algorithm configuration tool for finding the best configurations of a parameterised algorithm. One of its key ideas is \textit{racing}~\cite{maron1997racing}: using statistical tests to eliminate poor configurations early, avoiding wasting computational budget on less promising areas of the configuration space. \irace leverages this idea with an iterated procedure where each iteration is a \textit{race} among several configurations. Figure~\ref{fig:irace} illustrates \irace's tuning process. At the first iteration, a number of random configurations are generated, and a race started by evaluating all configurations on a subset of a given instance set, on a number of random seeds if the algorithm studied is stochastic, or a combination of both. A statistical test is applied to identify and eliminate the worst configurations. Evaluation proceeds with the remaining configurations and a statistical test is conducted again. This is repeated until only a few good configurations remain or when the budget for the current race has been used. The race is then finished and the surviving configurations are used to update a sampling model. In the next iteration, new configurations are generated based on the updated sampling model and a new race is started. Tuning terminates when a given number of evaluations is exhausted, and the best configuration(s) recorded are returned.

\subsection{\autoig's Instance Generation Process}
\label{sec:autoig}

\begin{lstlisting}[language=essence, caption={A fragment of an example for \racp problem}, label={lst:running_example_racp}, float]
% --- Fragment of MiniZinc model (succ: the immediate successors of tasks) --
array [int(1..n_tasks)] of set of int(1..n_tasks): succ;
% --- Fragment of generator model, in Essence ---
given n_tasks_t : int(1..60)        given s_density : int(1..5)
find succ: matrix indexed by [int(1..n_tasks_t)] of set of int(2..n_tasks_t)
such that sum([ |succ[t]| | t : int(1..n_tasks_t) ])/n_tasks_t = s_density
% --- Fragment of an example generator instance, in Essence ---
letting n_tasks_t = 6       letting s_density = 2
% --- Fragment of an example candidate instance, in MiniZinc ---
succ = [{2, 4, 5, 6}, {3, 4, 5}, {4, 5, 6}, {6}, {6}, {}];
\end{lstlisting}

We give an example of the instance generation process in \Cref{lst:running_example_racp}, based on \racp (see \Cref{sec:case_studies} for details). Fragments of a problem description model, a generator model, a generator instance, and a candidate instance are shown. 
In this example, a parameter (\code{succ}) of the problem description model (line 2) is written as a decision variable in the generator model (line 5). 
The creation of \code{succ} is controlled by tunable integer parameters of the generator model: \code{n\_tasks\_t} (equivalent to \code{n\_tasks} in the original problem description); and \code{s\_density}. Given an instance of the generator model sampled by \irace (line 8), a candidate instance (line 10) can be created by solving the generator instance. 

\autoig utilises \irace for searching in the configuration space of the generator model. 
The instance generation process starts with \irace creating a number of random \emph{generator configurations} (a configuration is an instance of the generator model, or in short, a \emph{generator instance}). Each configuration is then evaluated using the procedure described in Algorithm~\ref{alg:gen_configuration} and a penalty is given back to \irace for the statistical test. The tuning of \irace then proceeds as normal, interleaving using constraint solving to generate new instances and to evaluate them, and using feedback from the evaluation process to eliminate non-promising configurations and to update the sampling model.

\begin{algorithm}[t]
	\caption{An evaluation of a generator configuration} 
	\begin{algorithmic}[1]
	    \State \textbf{Input:} generator model $M$, generator instance $G$, solution history $H_G$
	    \State \textbf{Output}: penalty $p$
	    \State $r \leftarrow solve(M,G, H_G)$ \Comment{solve the generator instance $G$ using the \essence pipeline}
	    \If {$r$ is either UNSAT or timeout on \savilerow}
	        \State \Return $+\infty$ \Comment{return a very large penalty, \irace will discard $G$ immediately}
	    \EndIf
	    \If {$r$ is timeout on \minion}
	        \State \Return $1$
	    \EndIf
	    \State $I \leftarrow$ the instance generated by $r$
	    \State Add $I$ into $H_G$
	    \State $p \leftarrow$ Evaluate $I$ using either \Call{GRADED}{} or \Call{DISCRIMINATING}{} procedure 
	    \State \Return $p$
	\end{algorithmic} 
	\label{alg:gen_configuration}
\end{algorithm}

During each configuration evaluation, the generator instance $G$ is first solved via the \essence pipeline (line $3$ of Algorithm~\ref{alg:gen_configuration}), whose solving procedure includes two translation steps by the automated constraint modelling tool \conjure~\cite{akgun2011_AAAI,akgun2014_dontcare} and by \savilerow followed by a call to the constraint solver \minion~\cite{gent-minion-2006}. If $G$ is unsatisfiable or if it is too large to go through the pipeline, a very large penalty is returned so that \irace will remove the configuration from the current race immediately (line $5$). If $G$ is not solved by \minion within the current evaluation, a penalty of $1$ is returned. Otherwise, the new candidate instance $I$ is added to the solution history of $G$ to ensure that in the subsequent evaluations of this configuration, the same instance will not be generated again. Solution history is implemented via adding a negative constraint table into the \minion input of $G$, and this table is constantly updated every time $G$ is evaluated during the tuning. Finally, the candidate instance $I$ is evaluated using one of the two instance evaluation procedures described in Algorithm~\ref{alg:eval_graded} (for graded instance generation) or Algorithm~\ref{alg:eval_dis} (for discriminating instance generation), and the corresponding penalty is returned to \irace. Note that the default setting of \irace uses the Friedman test, a rank-based statistical test. This is also the setting used by \autoig, i.e., the magnitude of difference in the penalty values between evaluations is not taken into account, only the rankings between them matter.

\subsection{Evaluating Graded and Discriminating Instances}
\label{sec:instance_evaluation}

\autoig's instance generation process depends heavily on an effective way of evaluating the quality of candidate instances.
In this section, we describe the algorithms used for evaluating whether each candidate instance is graded or for measuring their discriminating power.
The algorithms given in this section are invoked in line $10$ of Algorithm~\ref{alg:gen_configuration}.

\begin{algorithm}[t]
	\caption{An evaluation of an instance using gradedness criteria} 
	\begin{algorithmic}[1]
	    \State \textbf{Input:} problem specification $P$, instance $I$, solver $S$, minimum solving time $t_{min}$, maximum solving time $t_{max}$, instance types $T$ (that we are interested in)
	    \State \textbf{Output}: penalty $p$
	    \Procedure{Graded}{$P,I,S,t_{min},t_{max},T$}
	    \State $r \leftarrow solve(P,I,S,t_{max})$ \Comment{solve $I$ using $S$ with time limit $t_{max}$, save results to $r$}
	    \If {$solving\_time(r) < t_{min}$ or $r$ is timeout}
	        \State \Return 0 \Comment{$I$ is either too easy or too difficult for $S$}
	    \EndIf
	    \If {$instance\_type(r) \not\in T$}
	        \State \Return 0 \Comment{$I$ is not the instance type we are interested in}
	    \EndIf
	    \State \Return -1
	    \EndProcedure
	\end{algorithmic} 
	\label{alg:eval_graded}
\end{algorithm}

To evaluate whether a candidate instance is graded, we employ \autoref{alg:eval_graded}. This algorithm has 6 inputs: a problem specification $P$ of the problem under study, an instance $I$ and a solver $S$ to be evaluated, the range of solving times ($t_{min}$ and $t_{max}$) for the instance to be considered graded for $S$ (to avoid instances that are too easy or too hard to solve), and the type of instances ($T$) that we are interested in (either satisfiable, unsatisfiable, or both).
The instance is first solved by $S$ (line $4$) (See Algorithm~\ref{alg:eval_graded}). Results of the solving ($r$) include the status of the solving process (timeout/UNSAT/SAT), and the returned solution $I$ (if status is SAT). In our experiments $S$ is called via the \minizinc toolchain. 
For complete solvers, we use the amount of time to solve the instance to completion (i.e., with a claim of optimality for optimisation problems, or with a feasible solution returned for decision problems or a claim of unsatisfiablity). For local search solvers such as \yuck, since a proof of optimality cannot be achieved for optimisation problems, we use an external complete solver (called the ``oracle'') to solve the instance to optimality (with a much longer time limit than $t_{max}$), and use that to measure the time until $S$ first finds the optimal solution. If the instance turns out to be too easy for $S$ or if the solving process times out (line $5$) or the instance type is not interesting to the users (line $7$), a penalty of $0$ is given back to \irace. Otherwise, the instance is considered graded and a negative penalty of $-1$ is returned. 

\begin{algorithm}[t]
	\caption{An evaluation of an instance using discriminating criteria} 
	\begin{algorithmic}[1]
	    \State \textbf{Input:} problem specification $P$, instance $I$, favoured solver $S_F$, base solver $S_B$, minimum solving time $t_{min}$ (for $B$ only), maximum solving time $t_{max}$, instance types $T$
	    \State \textbf{Output}: penalty $p$
	    \Procedure{Discriminating}{$P,I,S_F,S_B,t_{min},t_{max},T$}
	    \State $r_F \leftarrow solve(P,I,S_F,t_{max})$ \Comment{solve $I$ using $S_F$ with time limit $t_{max}$}
	    \State $r_B \leftarrow solve(P,I,S_B,t_{max})$ \Comment{solve $I$ using $S_B$ with time limit $t_{max}$}
	    \If {$r_F$ is timeout or $instance\_type(r_F) \not\in T$ or $solving\_time(r_{B}) < t_{min}$ }
	        \State \Return 0 \Comment{$I$ is either too difficult for $S_F$, or not the right instance type, or too easy for $S_B$}
	    \EndIf
	    \State $score_F, score_B \leftarrow$ \Call{MiniZinc\_Score}{$S_F,S_B,P,I$}
	    \If {$score_F=0$ and $score_B=0$}
	        \State \Return $0$
	    \EndIf
	    \State \Return $-score_F / score_B$\Comment{When \(score_B=0\), returns large negative number.}
	    \EndProcedure
	\end{algorithmic}
	\label{alg:eval_dis}
\end{algorithm}

\autoref{alg:eval_dis} is used for evaluating the discriminating power of an instance between two solvers. 
Each evaluation requires two input solvers: a \emph{favoured solver} $S_F$ and a \emph{base solver} $S_B$. We want to find instances that are easy to solve by $S_F$, while being difficult for $S_B$. The idea is to measure the performance of both solvers on the same instance, and search for instances that maximise the difference in performance. To avoid cases where the performance difference may be due to time measurement sensitivity, we impose a minimum solving time $t_{min}$ on the base solver $S_B$, i.e., the discriminating instances must be non-trivial to solve by $S_B$. 
Similar to the gradedness evaluation, \autoig also allows focusing on a particular instance type during the generation process. 

The evaluation of the discriminating property starts by applying $S_F$ and $S_B$ on the given instance (lines $4$ and $5$, Algorithm~\ref{alg:eval_dis}). If the instance does not satisfy our acceptance conditions (incorrect type, too easy for the base solver $S_B$ or unsolvable by the favoured solver $S_F$ (line $6$)) a penalty of $0$ is returned. Otherwise, we calculate the discriminating power of the instance and use it as feedback to \irace. The discriminating power is calculated as the ratio between the performance of the favoured solver and the base solver, and the aim of the tuning process is to maximise this ratio. To take into account both solving time and solution quality when evaluating the performance of a solver, we use the \emph{complete scoring} approach of the \minizinc competitions. After calculating the \minizinc scores of both solvers (line $8$), the discriminating score is calculated as the \minizinc score of $S_F$ divided by the \minizinc score of $S_B$ and the negation of that ratio is returned to \irace (line $11$). Note that when both \minizinc scores are equal to $0$, the discriminating score is set to $0$ (line $10$).

The \minizinc (complete) score for calculating the relative performance of two solvers on an instance can be found on the competition website (\url{https://www.minizinc.org/challenge2021/rules2021.html\#assessment}). For completeness, in the rest of this section we will describe this score calculation in detail. 

Given a solver $S$, a problem model $P$ and an instance $I$, the following information is collected for the calculation:
\Call{time}{$S,P,I$} -- the solving time of $S$ on $I$; \Call{solved}{$S,P,I$} -- whether a correct solution or a correct unsatisfiability result for $I$ is returned by $S$; \Call{quality}{$S,P,I$} -- the best objective value obtained by $S$; and \Call{optimal}{$S,P,I$} -- whether a claim of optimality is returned by $S$. Based on those information, the function \Call{IsBetter}{$A,B,P,I$} (Algorithm~\ref{alg:minizinc_is_better})  determines whether solver $A$ is clearly better than solver $B$ in terms of solution quality, for decision problems (line $4$) and for optimisation problems (lines $6$-$8$).

\begin{algorithm}[t]
	\caption{Check whether one solver performs better than another in terms of solution quality} 
	\begin{algorithmic}[1]
	    \State \textbf{Input:} solver $A$, solver $B$, problem model $P$, instance $I$
	    \Procedure{IsBetter}{$A,B,P,I$}
	        \If{$P$ is a decision problem}
	            \State \Return \Call{solved}{$A,P,I$} and not \Call{solved}{$B,P,I$}
	        \Else
	            \State \Return (\Call{solved}{$A,P,I$} and not \Call{solved}{$B,P,I$}) or 
	            \State (\Call{optimal}{$A,P,I$} and not \Call{optimal}{$B,P,I$}) or
	            \State (\Call{quality}{$A,P,I$} is better than \Call{quality}{$B,P,I$})
	        \EndIf
	    \EndProcedure
	\end{algorithmic} 
	\label{alg:minizinc_is_better}
\end{algorithm}

Finally, the \minizinc complete score when comparing two solvers on an instance $I$ is calculated in Algorithm~\ref{alg:minizinc_complete_scoring}. The calculation starts with checking whether one of the two solvers is better than the other in term of solution quality (lines $3$-$6$). If that is not the case, there are two possibilities. First, $I$ is solved by both solvers, and for optimisation problems, the same solution quality is achieved by both. In that case the normalised solving times are used as the scores. Second, both solvers fail to solve $I$, and in that case a score of $0$ is returned for both. Note that this is slightly different from the scoring used in the \minizinc competitions, where the scores of $1$ and $0$ are given to $A$ and $B$, respectively. This is because the final competition ranking is based on the Borda counting system, where the score is calculated for all pairs of solvers, including the same pair in the opposite order.

\begin{algorithm}[t]
	\caption{\minizinc score calculation between two solvers.} 
	\begin{algorithmic}[1]
	    \State \textbf{Input:} solver $A$, solver $B$, problem model $P$, instance $I$
	    \Procedure{MiniZinc\_Score}{$A,B,P,I$} 
	        \If{\Call{IsBetter}{$A,B,P,I$}}
	            \State $score_A \leftarrow 1$, $score_B \leftarrow 0$
	        \ElsIf{\Call{IsBetter}{$B,A,P,I$}}
	            \State $score_A \leftarrow 0$, $score_B \leftarrow 1$
	        \ElsIf{\Call{solved}{$A,B,P,I$}} 
	            \State $score_A \leftarrow $ \Call{time}{$B,P,I$}/(\Call{time}{$A,P,I$}+\Call{time}{$B,P,I$})
	            \State $score_B \leftarrow 1 - score_A$ 
	        \Else 
	            \State $score_A \leftarrow score_B \leftarrow 0$
	        \EndIf
	        \State \Return $score_A$ and $score_B$
	    \EndProcedure
	\end{algorithmic} 
	\label{alg:minizinc_complete_scoring}
\end{algorithm}

\section{Case Studies}
\label{sec:case_studies}

In this section we describe the five problems that are used to evaluate \autoig, and also the set of four solvers that are used in our experiments. 
The five problems being used in this study are taken from the latest \minizinc Challenges. They are chosen with the aim of covering a variety of different problem properties, including the existence of redundant and symmetry breaking constraints, the usage of different global constraints, and a range of problem domains. In this section, we give a brief overview of those problems and how their instance generation problems are modelled.

\textbf{Multi-Agent Collaborative Construction problem} (\macc)~\cite{lam2020exact}: This is a planning problem that involves constructing a building by placing blocks in a $3$D map using multiple identical agents. Ramps must be built to access the higher levels of the building. The objective is to minimise the makespan (primary) and the total cost (secondary). 

 In addition to the basic parameters of a \macc instance indicated in the problem specification (i.e., the number of agents, the time horizon and the map sizes), the instance generation process should include information about the building itself as this is likely to affect instance difficulty. Therefore, two parameters and related constraints are added to the generator model to represent the density of the building on the ground level and its average height.

\textbf{Carpet Cutting problem} (\carpetcutting):
The Carpet Cutting Problem \cite{schutt2011optimal} is a packing problem in which room and stair carpets composed of rectangular sections must be packed onto a carpet roll of fixed width and whose length must be minimised. The problem is complicated by the ability to rotate the carpets to aid in the packing process.

This problem requires substantial instance data, including the specification of the constituent rectangles of each carpet, their dimensions, and the permitted carpet rotations. There are several implicit constraints on this data that are not captured in the original \minizinc model and hence these must be injected into the instance generation process through our generator specification. In particular, the rectangles that comprise a carpet must not overlap and must form a contiguous shape, as well as have bounded sizes so as to avoid trivially unsatisfiable instances.

\textbf{Mario problem} (\mario): The Maximum Profit Subpath Problem is a routing problem that requires us to find a path in a graph where the path endpoints are given. This path is subject to two main constraints, where the sum of weights associated to arcs in the path is restricted (fuel consumption), while the sum of weights associated to nodes in the path has to be maximized (reward).

Regarding the instance generation process, in addition to the basic parameters, the amount of reward per node is represented as a non-negative integer array, while the non-negative cost for each arc is represented as a 2-dimensional matrix. There are a few implicit constraints not represented in the \minizinc model, where the initial and goal nodes are different and have 0 reward, and the cost matrix is symmetric on the diagonal.

\textbf{Resource Availability Cost Problem} (\racp): The Resource Availability Cost Problem~\cite{kreter2018mixed} is a scheduling problem with activities that are non-interruptible and have a fixed duration. The problem includes precedence constraints between pairs of activities \(i,j\) (that require activity \(i\) to be completed before activity \(j\) begins), arranged in a directed acyclic graph. There are a set of renewable resources, and each activity (when running) requires a given amount of each resource. All activities must be completed by a given deadline. Each resource has a cost per unit, and the objective is to minimise the peak costs of the resources. 

The durations of activities, unit costs of resources, and resource demands of activities are all matrices of integers without complex constraints. However, the precedence graph (represented as a set of successors for each activity) has implicit constraints that are not represented in the \minizinc model. Firstly, it must be acyclic, and we achieve this by mapping activities to numbered layers and allowing only edges from lower to higher-numbered layers. Secondly, we ensure that each activity has at least one predecessor and at least one successor (except the dummy first and last activities). 

\textbf{Discrete Lot Sizing problem} (\lotsizing): 
The Discrete Lot Sizing and Scheduling Problem \cite{houndji2014stockingcost,ullah2010literature} (CSPLib 58) requires us to find a production schedule for a set of orders, each with a due date within a planning horizon. There are various costs associated with production, such as setup, changeover and stocking costs, the sum of which must be minimised.

This problem requires substantial instance data including the type and due date of each order, and moreover a table of changeover costs between orders. There are a number of implicit constraints on this data, including a dummy order type 0 which incurs 0 cost to change to/from, and the fact that the changeover costs for the remaining order must obey the triangle inequality. Again, these are not captured in the original \minizinc model and hence must be injected into the instance generation process through our generator specification.


We investigate the performance of four solvers, also taken from the \minizinc challenges, on the problems described above using our framework. They are chosen such that a variety of solving techniques and different competition rankings are included. The solvers are: \ortools~\cite{ortools} (version 9.2) -- a systematic solver from Google that combines CP, SAT, and linear programming techniques; \picat~\cite{zhou2017optimizing} -- a SAT compiler for the multi-paradigm programming language Picat which uses \textsc{kissat}~\cite{kissat2021} as the underlying SAT solver; \chuffed~\cite{chuffed} (version 0.10.4) -- a clause learning CP solver which was not a participant of the challenges but was used in the score calculation process to rank participating solvers; and \yuck~\cite{yuck} (version 20210501) -- a constraint-based local search solver.

\ortools has consistently won the last several competitions and \picat has received multiple silver medals. \yuck is the winning solver in the Local Search category of the 2020 and 2021 competitions. However, its ranking was generally low when compared to \ortools and \picat. In particular, based on the competition data, it was completely dominated by \ortools on the five problems considered.

\section{Experimental Setup}
\label{sec:experimental_setup}

The first set of experiments are on generating graded instances. For each problem, we first generate graded instances for each solver via an \autoig experiment with a budget of $2,000$ runs. Note that a run is an evaluation of a generator configuration. The gradedness criteria is defined as being solvable by the given solver with the time ranging from $10$ seconds (to avoid trivial instances) to $20$ minutes (the time limit used by the \minizinc Challenge). Following the competition approach, \minizinc translation time is included in the total time measured. Since \yuck is a local search solver, we use \ortools (with a budget of $1$ hour) for checking whether a solution returned by \yuck is optimal. After all graded instances are collected, we then randomly select $50$ graded instances from each experiment to get a combined benchmark instance set for each problem. Finally, we evaluate the performance of all four solvers on the combined instance set.

The second set of experiments are on generating discriminating instances. Since \ortools has consistently shown very strong performance on the competition data, the main aim of these experiments is to see whether we can find instances where \ortools is performing worse than the other two participating solvers being considered. We do this without loss of generality: our discriminating instance generation procedure can be applied to any pair of solvers. We compare two solvers (\picat and \yuck) against \ortools. For each solver we conduct two separate \autoig experiments, one where we search for instances that are solved more quickly by \ortools and one for the opposite case. The same \autoig budget and memory limit as in graded experiments are used. To avoid instances where the difference between the performance of two solvers is due to fluctuations in running time measurement, a minimum requirement of $10$ seconds is imposed on the solving time of the base solver, i.e. instances that can be trivially solved by the base solver are discarded.

All experiments were performed on a computing node of a High Performance Computing cluster. Each node is equipped with two $2.1$ GHz, $18$-core Intel Xeon processors and $256$ GB RAM. Each solver except \yuck is given a memory limit of $8$GB via the \runsolver tool~\cite{roussel2011controlling}. For \yuck, the memory limit is controlled directly via the Java Runtime Environment (JRE). For solving the generator models, time limits of $5$ and $10$ minutes are given to \savilerow and \minion, respectively. In this work, we focus on the Free Track of the competitions. Therefore, all solvers are called via the \minizinc toolchain with a single core and with the free search option being passed to the solver. Although \autoig supports focusing on generating either only SAT or only UNSAT instances, in this work we allow both types of instances to be generated.

\section{Results on graded instances}
\label{sec:results_graded}

First we describe the sets of graded instances produced by \autoig for the five problems (\Cref{sec:results_graded_instancegen}) and discuss insights obtained from analysing the results. Then in \Cref{sec:results_graded_comparison} we combine the sets of graded instances for each problem, and re-evaluate the four solvers using the combined sets of instances, showing substantially different relative performance in some cases compared with the competition instances.

\subsection{Graded instance generation}
\label{sec:results_graded_instancegen}

\begin{figure}[t]
    \centering
    \includegraphics[width=0.6\linewidth,trim=0cm 1.5cm 0cm 1cm,clip]{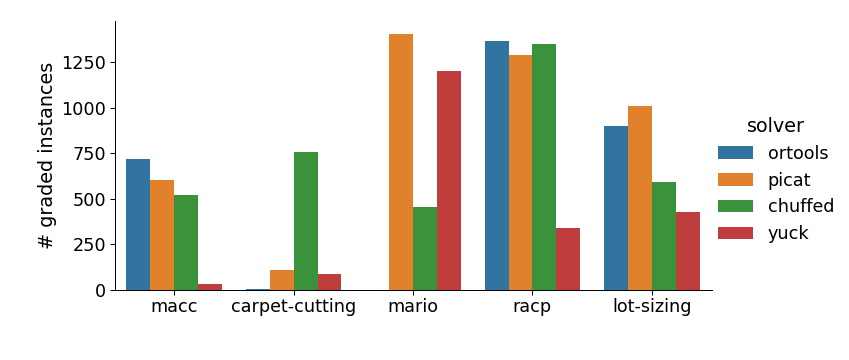}
    \caption{Number of graded instances generated.}
    \label{fig:number_of_graded_instances}
\end{figure}

For each problem, Figure~\ref{fig:number_of_graded_instances} shows the number of graded instances obtained per solver within the given budget. While we can achieve more than a few hundred graded instances in most cases, there are cases where we are only able to generate a small number of instances. For example, with \ortools on \carpetcutting and \mario, we generate only $4$ and $1$ graded instances, respectively. In addition, the numbers are fairly small for \yuck on \macc and \carpetcutting. There is a large variation in the number of graded instances we are able to generate for different problems and solvers (shown in \autoref{fig:number_of_graded_instances}).

\begin{figure}[t]
    \centering
    \includegraphics[width=\linewidth]{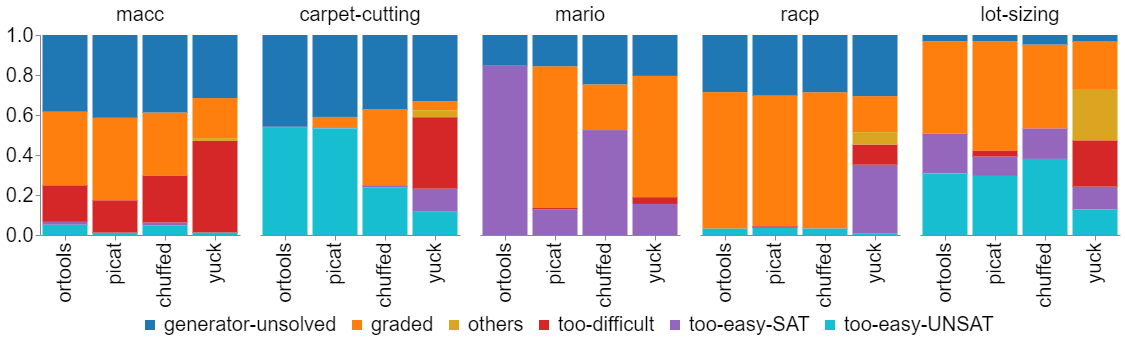}
    \caption{Frequency of all run statuses, including \texttt{generator-unsolved} (generator instance is UNSAT or unsolvable); \texttt{graded} (a graded instance is obtained); \texttt{too-difficult} (the candidate instance is unsolvable by the considered solver within the time limit); \texttt{too-easy-SAT} and \texttt{too-easy-UNSAT} (the candidate instance is too easy, i.e., solved within less than $10$ seconds); and \texttt{others} (the considered solver fails due to unexpected errors such as incorrect returned answers). }
    \label{fig:graded_details}
\end{figure}

The differences in the number of graded instances returned by each experiment suggest that the performance of the solvers varies significantly when solving instances drawn from the same instance space. In order to better understand the performance distribution of each solver we  investigate the details of the search space of \autoig. More specifically, we check the status of each configuration evaluation run and measure their frequency, as detailed in \autoref{fig:graded_details}. For \ortools on \carpetcutting and \mario, only a small number of graded instances are found, but this same outcome has entirely different causes.  For \carpetcutting, almost half of the runs are with unsolvable generator configurations, and for the rest the candidate instances are mostly trivially proved unsatisfiable by \ortools{}. For \mario, the majority of the runs produce instances that are trivially satisfiable. Once we understand the underlying reason for the lack of graded instances, we can rectify each of these shortcomings: for \carpetcutting, expert knowledge on the problem may be added as constraints to the generator model to avoid trivially unsatisfiable instances, while for \mario, the current instance space may be too easy for \ortools and we may want to increase the upper bounds of some of the generator parameters. On the other hand, the situation is completely different for \yuck: the small number of graded instances obtained for \macc and \carpetcutting is largely due to the fact that the majority of instances generated are too difficult to solve.

\begin{figure}[t]
    \centering
    \includegraphics[width=0.9\linewidth,trim=0cm 2cm 0cm 2cm,clip]{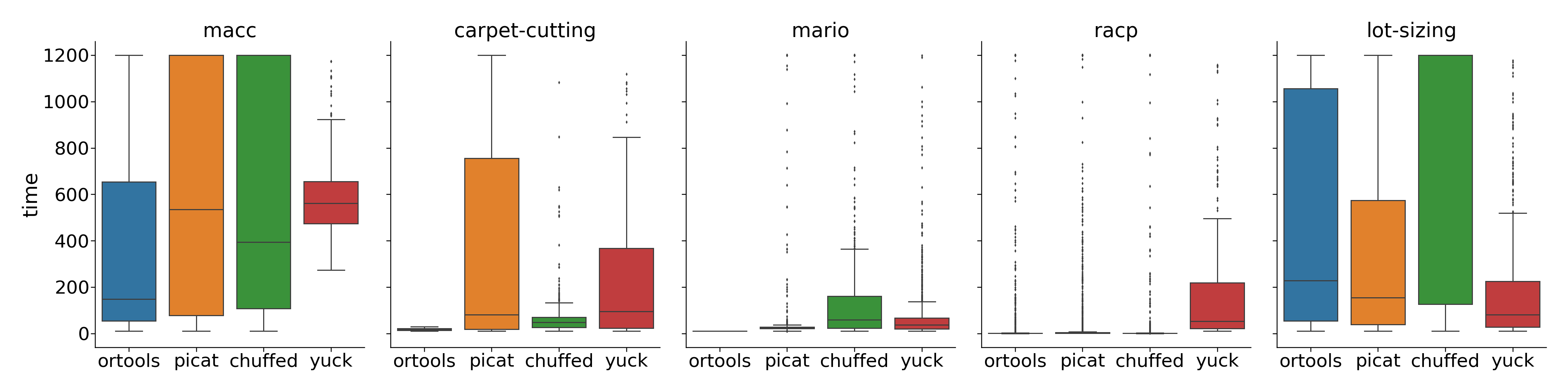}
    \caption{Solving time of graded instances generated for each pair of problems and solvers. Note that the instances presented here are the graded instances found for each solver independently. The performance of these solvers on the combined set of graded instances can be seen in \autoref{fig:graded_comparison}.}
    \label{fig:graded_time_distribution}
\end{figure}

In addition to the run statuses, the distribution of solving time of graded instances also gives us interesting insights into the performance of different solvers, as illustrated in \autoref{fig:graded_time_distribution}. Notably, many graded instances for \mario and \racp are close to the lower bound of graded instances; this is true for all solvers. Nevertheless, \autoig is able to find challenging graded instances, which can take several hundred seconds to solve, for all solvers on those two problems (except for \ortools on \mario). For \carpetcutting, \ortools and \chuffed can solve most graded instances quickly, while \picat and \yuck take more time in general. Finally, for \macc and \lotsizing, the solving time distributions of all four solvers are more well-spread, indicating a good diversity of difficulties among the generated graded instances.

Note that for the majority of graded instances generated, the \minizinc flattening times are generally marginal compared to the time taken to solve them. This indicates that the more difficult graded instances are actually challenging for the solvers themselves, and can be useful for solver developers to improve their solver performance. 

\subsection{Comparison of Solver Performance on Graded Instances\label{sec:results_graded_comparison}}

We combine all graded instances to construct a diverse set of instances for each problem. We then evaluate all four solvers on the combined set and rank them using the Borda (complete) scoring method of the \minizinc Challenge (\url{https://www.minizinc.org/challenge2021/rules2021.html}).
More specifically, for each problem, $50$ graded instances are uniformly sampled from the set of graded instances for each solver. In cases where there are less than $50$ graded instances available, we just take them all. For comparison, we also evaluate those solvers on the instances used in the competition. There are $5$-$10$ instances per problem, as some problems are re-used over two different competitions.

\begin{figure}[t]
    \centering
    \includegraphics[width=0.45\linewidth,trim=0cm 0.25cm 0cm 0.25cm,clip]{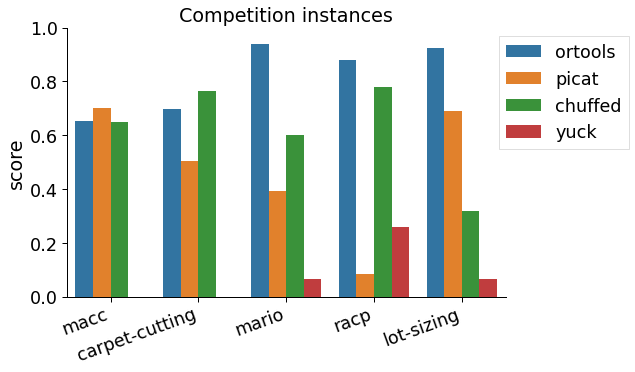}
    \includegraphics[width=0.38\linewidth,trim=0cm 0.25cm 0cm 0.25cm,clip]{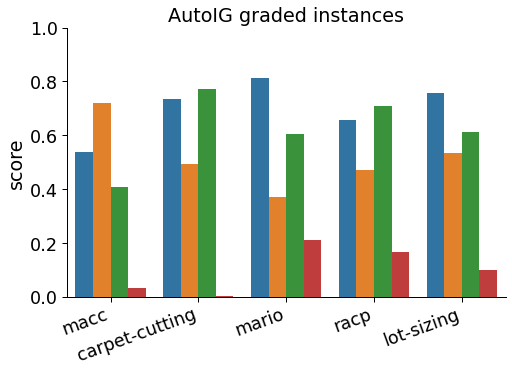}
    \caption{\minizinc Borda (complete) scores of each solver on the \minizinc Challenges instance set (left) and on the combined graded instance set generated by \autoig (right).}
    \label{fig:graded_comparison}
\end{figure}

Figure~\ref{fig:graded_comparison} shows the scores on the competition instances (left) and on the combined graded instances generated by \autoig (right). There are similarities between results on the two sets of instances. Performance of \ortools and \chuffed  remain strong in most cases, followed by \picat. For \macc, \carpetcutting and \mario, the overall rankings of the four solvers on both groups are almost the same. However, results on the graded set do show certain changes in relative performance of all solvers. For example, the scores of \yuck on the graded instances are no longer zero for \macc and \carpetcutting, and the score for \mario increases noticeably. This indicates that \yuck is actually not completely dominated by all other solvers on those three problems as suggested by the competition data. For \racp, the ranking has changed significantly: \ortools swaps places with \chuffed, and \picat swaps places with \yuck. For \lotsizing, \picat is no longer ranked higher than \chuffed.

Thanks to the solution checking process being integrated into each evaluation, we also found a number of cases from the combined graded sets where incorrect answers are returned, which can be of separate interest to the solver developers. There were $41$ (out of $183$) \macc instances and $90$ (out of $154$) \carpetcutting instances (from the subset of graded instances generated for other solvers) where \yuck reports objective values of infeasible solutions. 

Generating a larger number of graded instances for each solver and analysing them using the presented methods gives more information in comparison to a typical competition's result, which would be a ranking of the solvers. In \autoref{sec:results_discriminating} we apply the discriminating instance generation feature of \autoig to gain even more insight into solver performance.

\section{Results on Discriminating Instances}
\label{sec:results_discriminating}

Results on \minizinc competition data indicate that \ortools is a very strongly performing solver on the $5$ problems considered. It completely dominates \yuck, i.e., \yuck gets zero score on all competition instances when compared directly to \ortools. \ortools also wins over \picat on all instances of \mario and \racp, on $9$ out of $10$ instances of \lotsizing, and on $8$ out of $10$ instances of \carpetcutting. However, detailed results obtained from the evaluation on graded instances suggest that this may not always be the case. For example, there are $31$ instances evaluated on \racp where \picat performs better than \ortools, and $58$ \macc instances where \yuck performs better. In this section, we use the discriminating instance generation feature of \autoig to get more insights into these cases.

\begin{figure}[t]
    \centering
    \includegraphics[width=0.5\linewidth,trim=0.7cm 0.7cm 0.7cm 0.7cm,clip]{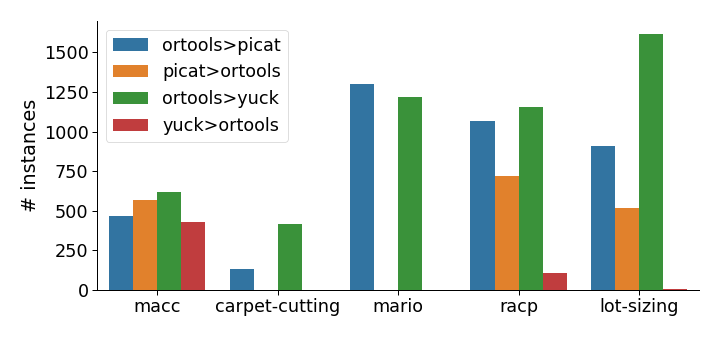}
    \caption{Number of discriminating instances generated per \emph{favoured} and \emph{base} solver pair.}
    \label{fig:number_of_dis_instances}
\end{figure}

Figure~\ref{fig:number_of_dis_instances} shows the number of discriminating instances generated for the two pairs of solvers. In the experiments on \ortools versus \yuck, \autoig found $431$ \macc instances and $110$ \racp instances where \yuck gets a better score than \ortools, which indicates that \yuck is not completely dominated by \ortools on these two problems. On the other hand, for \carpetcutting and \mario, results suggest that \yuck may indeed be entirely dominated by \ortools, as no instances were found in the experiments that favour \yuck. Furthermore, for \lotsizing, only $3$ discriminating instances favouring \yuck are found. In the experiment on \ortools versus \picat, \ortools shows domination on both \carpetcutting (only $2$ instances where \picat is better than \ortools were found) and \mario (no instances favouring \picat was found). On the other three problems, there are a good number of discriminating instances in both directions.

\begin{figure}[t]
    \centering
    \includegraphics[width=0.9\linewidth,trim=2cm 2cm 2cm 2cm,clip]{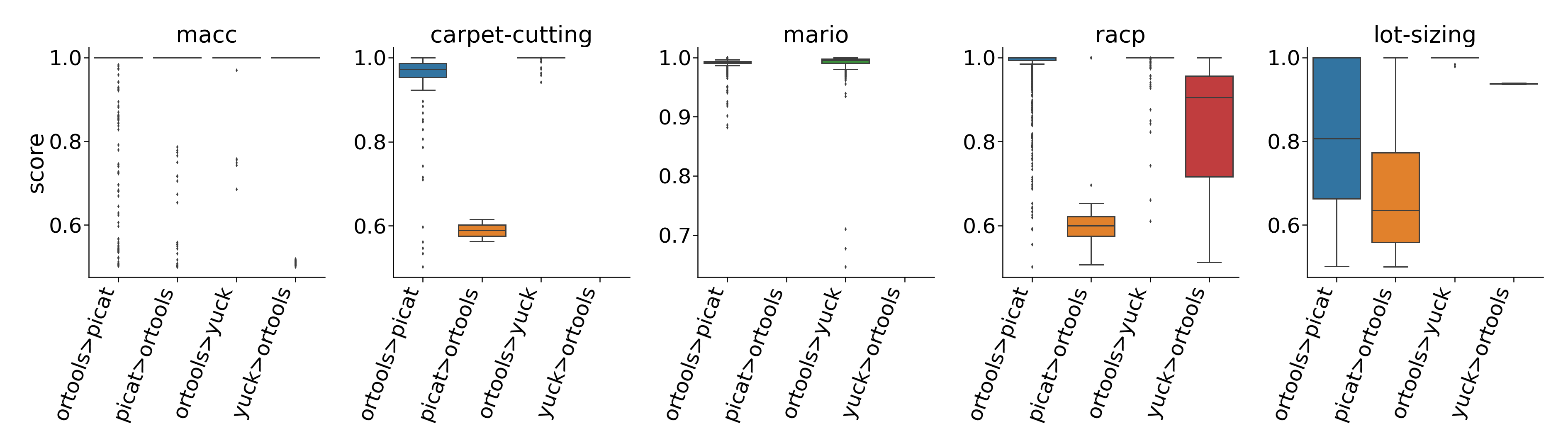}
    \caption{Distribution of scores (of the winning solver) on discriminating instances generated.}
    \label{fig:dis_score_distribution}
\end{figure}

The number of discriminating instances tell us if winning instances for a solver can be found, but it does not show the magnitude of the difference in performance. We can get additional insights into comparative performance of the solvers by looking into the detailed scores of the winning solver on discriminating instances for each experiment. As shown in Figure~\ref{fig:dis_score_distribution}, for \macc, the median lines indicate that for all four cases, several discriminating instances found have the highest ``discriminating power'', i.e., the winning solver gets the maximum score of $1$ (the other solver, in turn, gets zero score). This type of instance is probably the most interesting for understanding the shortcomings of a particular solver. For \carpetcutting, on the only $2$ discriminating instances where \picat has better score than \ortools, the score distribution of the corresponding experiment (\picat{}>\ortools{}) suggests that \ortools performance is not much worse. This suggests that \ortools indeed dominates \picat on this problem. A similar conclusion can be reached for \yuck, i.e., it is clear that \ortools is really the dominating solver on \carpetcutting since the magnitude of the performance difference is very small even for the instances where \yuck is faster. Similarly, for \mario, \ortools very clearly dominates in comparison to \picat and \yuck, as indicated by the discriminating score distributions. This is in line with what was observed in the previous section's results on the same problem. 

Interestingly, for \racp, although the number of discriminating instances of \picat{}>\ortools{} is larger than of \yuck{}>\ortools{} as shown in Figure~\ref{fig:number_of_dis_instances}, the magnitude of the performance difference of instances found for \yuck is generally much higher. This observation gives a new insight that has not been revealed in all previous experiments on gradedness: even though the performance of \yuck is dominated by other solvers in general (i.e., it is ranked lower) and it has a smaller number of discriminating instances favouring it, the magnitude of the performance difference is very large for these instances. This means there exists a subset of the \racp instances where \yuck's performance is much better than \ortools, while this does not seem to be the case for \picat. 

The insights provided by discriminating instances could be useful in constructing a robust portfolio of solvers for a given problem. For example, on \racp{}, \yuck{} is the weakest solver by a wide margin on the graded instances (see \autoref{fig:graded_comparison}) and second-weakest on competition instances. On the graded instances, \picat performs considerably better than \yuck. However, the results with discriminating instances show that \yuck{} would be a good candidate to add to a portfolio (alongside \ortools) whereas \picat may not be. 

\section{Conclusions and Future Work}
\label{sec:conclusion}

Assessing the performance of solving methods via benchmark problems is fundamental to CP research. However, its utility is limited by the availability of problem instances that are of suitable difficulty, and diverse (not inadvertently favouring one solver over another). We have shown that our system \autoig can generate large numbers of informative benchmark instances graded for difficulty for a single solver, or that can discriminate between two solvers (favouring one or the other).  The only manual part of the \autoig process is to capture (in a generator model) any implicit constraints on the instances data.

The essential task of benchmarking is to compare multiple solvers and rank them. As illustrated in our experiments, \autoig{} can be used to generate graded instances for each solver independently, and these can then be combined into one set of instances, providing confidence that the generation process does not favour one solver or class of solvers. Furthermore, we have shown that automatically generated instances can provide more detailed insights than just a ranking. Instances generated by \autoig{} can reveal cases where a solver is weak or even faulty, providing valuable information to solver developers. Finally, discriminating instances can reveal parts of the instance space where a generally weak solver performs well relative to others, and therefore could be useful as part of a portfolio.  

There are various directions for future improvement. First, the diversity of instances found during search can be taken into account to increase the quality of the final instance set. This would require a definition of diversity, which could be based on problem-specific instance features or on general constraint programming features such as the \textsc{fzn2feat} features~\cite{amadini2014enhanced}.  
Secondly, similar to the series of work on Instance Space Analysis (e.g.~\cite{munoz2018instance,kletzander2021instance,de2021algorithm}), a detailed visualisation of the instance space based on performance data collected from the tuning and evaluation process of \autoig would provide further insights into performance of the solvers under study. Again, instance features would be needed for such analysis.


\newpage

\bibliographystyle{plainurl}
\bibliography{references}

\end{document}